\pdfoutput=1
\documentclass[letterpaper]{article}
\usepackage[preprint]{aaai2027}
\usepackage[hyphens]{url}
\usepackage{graphicx}
\usepackage{natbib}
\usepackage{caption}
\usepackage{subcaption}
\usepackage{booktabs}
\usepackage{amsmath}
\usepackage{amssymb}
\usepackage{array}
\usepackage{multirow}

\title{SKILLER: Language-Level Reinforcement Learning for Reusable Skill Extraction in Small Language Models}

\author{
Chenhao Dang\textsuperscript{\rm 1,\rm 2}\equalcontrib,
Siyuan Xiong\textsuperscript{\rm 3}\equalcontrib,
Conghui He\textsuperscript{\rm 2}\corresponding,
Weijia Li\textsuperscript{\rm 2,\rm 4}\corresponding
}
\affiliations{
\textsuperscript{\rm 1}Shanghai Jiao Tong University,
\textsuperscript{\rm 2}Shanghai Artificial Intelligence Laboratory,
\textsuperscript{\rm 3}Harbin Institute of Technology, Shenzhen,
\textsuperscript{\rm 4}Tsinghua Shenzhen International Graduate School, Tsinghua University,
\\dangchenhao@pjlab.org.cn, xiongsiyuan@stu.hit.edu.cn,
heconghui@pjlab.org.cn, liweijia@sz.tsinghua.edu.cn
}

\newcommand{\method}{SKILLER}
\newcommand{\best}[1]{\textbf{#1}}
\newcommand{\na}{--}

\begin{document}
\maketitle

\begin{abstract}
Agent skills represent a standardized format for packaging procedural knowledge and domain expertise, serving within agent harness systems as an essential mechanism to continually constrain a language model's behavior space for repeatable, high-quality task execution. However, because strong closed-source models entail high inference costs, current popular agent harnesses, such as Codex and OpenClaw, remain prohibitively expensive when deploying these skills to accomplish real-world tasks. The rapid capability enhancement of open-source models deployable on consumer-grade GPUs presents a compelling opportunity to drastically reduce these costs by leveraging skill-based behavioral constraints. Nevertheless, automatically generating effective skills tailored specifically for such compact models remains a significant practical challenge. To address this, we propose SKILLER, a natural-language-driven reinforcement learning framework designed to automatically generate executor-specific skills for small models, which employs a strong model as the actor and critic, treats the small-model agent system as the environment, and propagates all reinforcement learning signals entirely via natural language. Extensive experimental evaluations across five relevant benchmarks using Qwen3.5-9B and Qwen3.5-4B demonstrate that SKILLER outperforms three open-source and one closed-source skill generation or evolution methods, achieving absolute gains ranging from 4.3 to 20.4 percentage points for the 9B model and 1.8 to 13.3 points for the 4B model, while remarkably matching the performance of strong closed-source models on single-skill tasks in SkillsBench. The project is available at \url{https://github.com/DANG-ai/SKILLER}.
\end{abstract}

\section{Introduction}

In the rapidly evolving landscape of autonomous agents, agent skills have emerged as a foundational primitive \citep{zhang2025equippingagents,ling2026agentskills}. Conceptually aligned with recent frameworks formalized by AI research organizations such as Anthropic, an agent skill is not merely a prompt, but a standardized format for packaging procedural knowledge, tool-use conventions, and domain expertise \citep{anthropic2026agentskillsdocs,li2026skillsbench}. Within advanced agent harness systems, these skills serve as an essential mechanism to continually constrain the behavior space of Large Vision-Language Models (LVLMs), thereby ensuring that complex tasks are executed in a repeatable and high-quality manner. However, achieving this reliability traditionally relies on strong, closed-source frontier models. Because these models entail exorbitant inference costs, current popular agent harnesses, such as Codex, Claude Code, OpenCode, and OpenClaw, remain prohibitively expensive when deploying skills to accomplish real-world tasks at scale \citep{anthropic2026claudecode,openai2026codexweb,opencode2026docs,openclaw2026skills}.

\begin{figure}[t]
  \centering
  \includegraphics[width=\columnwidth]{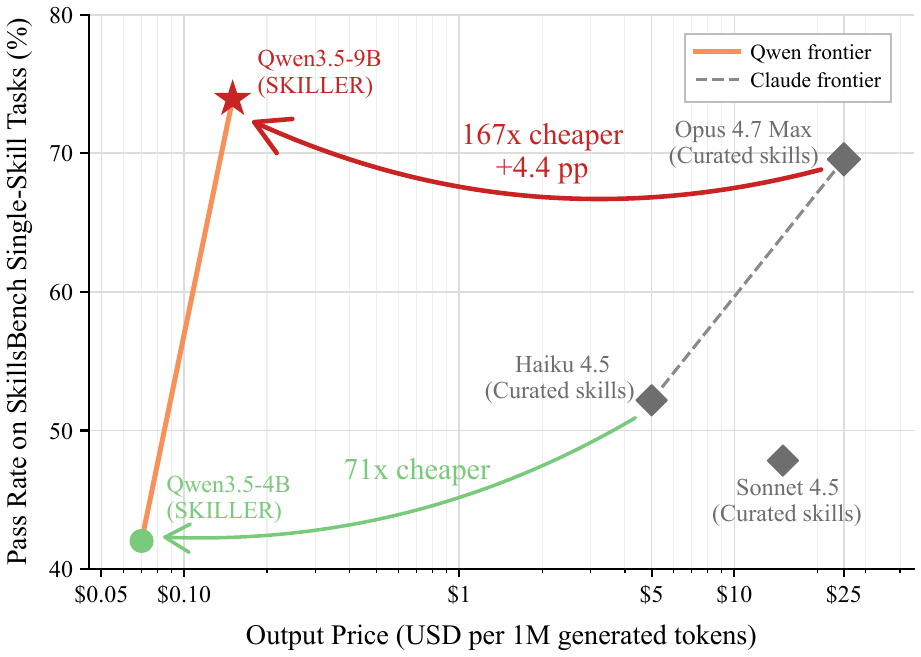}
  \caption{Cost--performance on single-skill tasks of SkillsBench. Skills generated by \method{} enable Qwen3.5-based agent loops to approach the frontier at much lower output-token prices. On specific tasks, Qwen3.5-9B with \method{} skills even outperforms closed-source models using curated skills, suggesting that well-matched agent skills can make lower-cost models more effective than more expensive alternatives, as in the case of Haiku 4.5 outperforming Sonnet 4.5 \citep{li2026skillsbench}.}
  \label{fig:teaser}
\end{figure}

A promising alternative to this cost bottleneck lies in the rapid capability enhancement of open-source compact models deployable on consumer-grade GPUs, such as the Qwen3.5, Qwen3.6, and Gemma 4 series \citep{qwen35blog,googledeepmind2026gemma4modelcard}. When augmented with specific, well-defined skill constraints, these compact models exhibit surprising proficiency \citep{xu2026slmskills,li2026skillsbench}. For simple or highly repetitive real-world tasks, guiding these small-scale LVLMs with strict procedural boundaries not only can maintain high success rates but also drastically reduce operational costs \citep{xu2026slmskills,li2026skillsbench}. As illustrated in Figure~\ref{fig:teaser}, when equipped with appropriate skills, compact models like Qwen3.5-9B and Qwen3.5-4B can achieve task-specific performance that rivals that of their massive, closed-source counterparts, unlocking a highly cost-effective paradigm for agentic deployment.

Despite this immense potential, a critical model-mismatch problem hinders direct deployment. Skills crafted for strong frontier models do not natively transfer to small-scale LVLMs \citep{huang2026rawexperience,liu2026wildskills}. The behavioral constraints, implicit reasoning steps, and error-recovery assumptions that successfully guide a massive model are often completely ineffective for a compact model. A small-scale LVLM might easily hallucinate arguments, skip essential verification steps, or become derailed by overly complex instructions that feature multiple branching paths. Consequently, directly porting existing high-end skills to these compact models frequently leads to catastrophic task failure. How to automatically and reliably generate effective skills tailored specifically to the unique behavioral spaces of such small-scale LVLMs remains a practical challenge.

To overcome this challenge, we propose \method{}, a novel natural-language-driven reinforcement learning framework designed to automatically generate and optimize executor-specific skills for small-scale LVLMs. Unlike traditional reinforcement learning paradigms that update neural weights, \method{} treats the textual skill itself as the optimizable policy. The framework employs a state-of-the-art frontier model, such as GPT-5.5 or Claude opus 4.8, to serve as the actor and critic. The environment within this reinforcement learning formulation operates as an agent loop driven by the target open-source, small-scale LVLM. This environment functions under a progressive skill disclosure mechanism that prevents the compact model from being overwhelmed by long textual contexts. Crucially, all reinforcement learning signals in \method{}, including states, diagnostic rewards, and policy update actions, are propagated entirely via structured natural language, bridging the gap between strong-model reasoning and small-scale LVLM execution.

We comprehensively evaluate \method{} using Qwen3.5-9B and Qwen3.5-4B across a diverse suite of benchmarks. Our evaluation encompasses four general-purpose benchmarks, namely SkillsBench \citep{li2026skillsbench}, SkillLearnBench \citep{zhong2026skilllearnbench}, SWE-Skills-Bench \citep{han2026sweskills}, and GAIA \citep{mialon2023gaia}, alongside one specialized domain benchmark designated as EarthBench \citep{feng2025earthagent}. Furthermore, we benchmark our framework against three open-source skill generation and evolution methods, specifically EvoSkill \citep{alzubi2026evoskill}, AutoSkill \citep{yang2026autoskill}, and SkillX \citep{wang2026skillx}, as well as one closed-source skill-generation baseline known as Manus \citep{manus2026skills}. Empirical results demonstrate that the skills generated by \method{} yield substantial and consistent performance improvements within the agent loops of both the 9B and 4B compact models. These findings confirm that language-level reinforcement learning effectively unlocks the autonomous capabilities of cost-efficient small-scale LVLMs.Our main contributions of this work are summarized as follows.

\begin{itemize}
\renewcommand{\labelitemi}{$\bullet$}
\item \textbf{Novel Skill Generation Framework.} We introduce \method{}, a natural-language-driven reinforcement learning framework designed to automatically generate executor-specific skills for small-scale LVLMs. Treating textual skills as optimizable policies, our approach leverages a strong frontier model to serve as the actor and critic within the interactive environment of the target compact model's agent loop.
\item \textbf{Language-Level Policy Iteration.} We formulate a language-level policy iteration mechanism that propagates all reinforcement learning signals, including states, diagnostic rewards, and policy update actions, entirely via structured natural language. This feedback loop resolves the model-mismatch gap by diagnosing and repairing the unique failure modes of compact models, generating tailored constraints without any neural weight updates.
\item \textbf{Comprehensive Empirical Validation.} We extensively evaluate \method{} across five diverse benchmarks using the Qwen3.5-9B and Qwen3.5-4B compact models, consistently outperforming four baseline skill evolution methods. The generated skills empower these small-scale LVLMs to rival the task-specific performance of massive closed-source counterparts, establishing a highly cost-effective paradigm for agentic deployment.
\end{itemize}

\section{Related Work}

\paragraph{Agent skills.} Agent systems frequently interleave reasoning with tool execution \citep{yao2023react,schick2023toolformer}, and agent skills package these procedural choices into reusable memory artifacts. The evaluation of these skills across diverse benchmarks highlights both their utility and their uneven transferability. For instance, curated skills can improve execution success \citep{li2026skillsbench}, whereas mismatched or partially relevant skills often degrade performance \citep{han2026sweskills,liu2026wildskills}. Furthermore, comprehensive benchmarks systematically evaluate continual skill generation \citep{zhong2026skilllearnbench}, tool composition abilities \citep{chen2026skillcraft}, and lifelong library maintenance \citep{zhang2026skillflow}. Because well-crafted skills provide strict procedural boundaries that restrict the action space, they are exceptionally well-suited for guiding small-scale LVLMs to complete tasks efficiently in specific structured scenarios.

\begin{figure*}[t]
  \centering
  \includegraphics[width=1.0\linewidth]{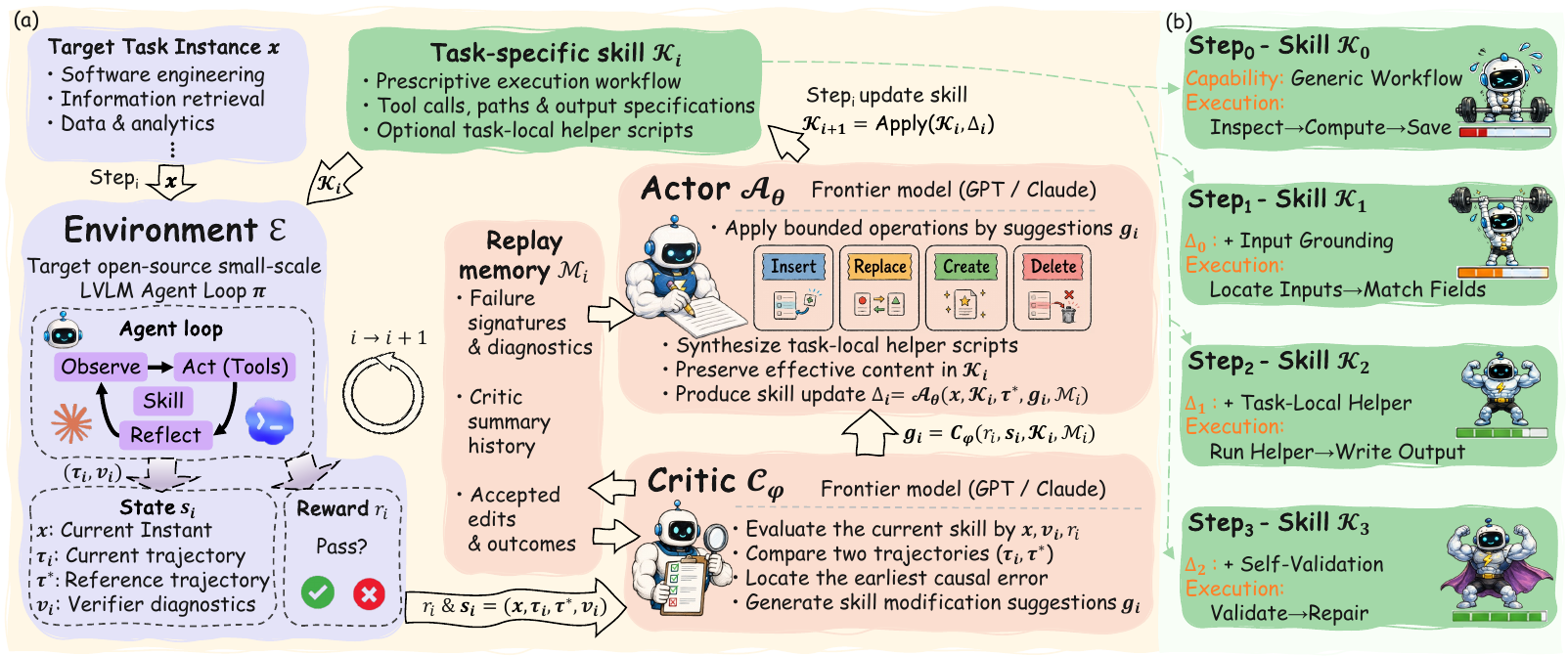}
  \caption{Overview of \method{}. \textbf{(a)} Automated generation and iterative refinement of task-specific skills tailored for compact models. At each optimization step, the framework takes a target task instance and the current skill, treating the small-scale LVLM agent loop as an interactive environment to produce a structured state quadruple and a scalar reward. Driven by a frontier model, the critic analyzes the current state alongside historical records from the replay memory to evaluate execution outcomes, isolate causal errors, and formulate modification suggestions before saving these diagnostics back into memory. The actor then executes bounded editing operations to update the skill content. Crucially, all information transfer across this optimization loop is conducted entirely via natural language. \textbf{(b)} The progressive performance enhancement of the small-scale LVLM agent loop as the underlying skill iteratively evolves and strengthens across successive optimization steps.}
  \label{fig:framework}
\end{figure*}

\paragraph{Skill acquisition and evolution.} Prior literature explores skill acquisition from diverse sources, including public repositories \citep{bi2026mining}, large-scale networks \citep{liang2026skillnet}, interaction histories \citep{yang2026autoskill}, collective trajectories \citep{ma2026skillclaw}, and successful agent rollouts \citep{alzubi2026evoskill,wang2026skillx}. Advanced frameworks facilitate this evolution by maintaining executable libraries \citep{wang2023voyager}, storing verbal feedback \citep{shinn2023reflexion}, distilling local lessons \citep{ni2026trace2skill}, managing dual expertise forms \citep{qiu2026autorefine}, and utilizing co-evolutionary verification \citep{zhang2026coevo}. Recent methods also integrate skills with reinforcement learning by modifying agent policies, optimizing skill banks, or inducing programmatic constraints \citep{wang2025sage,xia2026skillrl,mi2026skillpro,wang2025programmatic}. However, these existing generation and evolution techniques are predominantly optimized for massive frontier models, and there are currently no skill generation methods specifically tailored to address the unique constraints and execution paradigms of compact models.

\section{Method}
\label{sec:method}

\method{} is a natural-language policy optimization framework that translates frontier-model reasoning into task-specific skills for small-scale LVLMs, without updating the parameters of the compact model itself. As illustrated in Figure~\ref{fig:framework}, \method{} treats the textual skill as the optimization variable, the small-scale LVLM agent loop as an interactive environment, and the official benchmark verifier as the source of task reward. Driven by a frontier model, a critic module converts execution evidence into causal and localized feedback, while an actor module implements this feedback through bounded edits to the skill artifact. A structured replay memory preserves failure signatures, critic diagnoses, and previously effective modifications across sequential optimization steps. This formulation concentrates computational expense at skill-construction time, yields an interpretable and directly deployable policy artifact, and propagates all diagnostic feedback and editing instructions entirely through natural language rather than gradient updates.

\subsection{Problem Definition}
\label{sec:problem-definition}

Let $\mathbf{x}\in\mathcal{X}$ denote a target task instance, including its instruction, runtime inputs, available tools, and output contract. A frozen compact model $\pi$ interacts with a task environment $\mathcal{E}$ under a task-specific skill $\mathcal{K}_i$ at optimization step $i$. The skill conditions the compact model's action distribution, inducing the effective policy
\begin{equation}
    \pi_{\mathcal{K}_i}(a_t\mid h_t,\mathbf{x})
    \triangleq
    \pi(a_t\mid h_t,\mathbf{x},\mathcal{K}_i),
\end{equation}
where $h_t$ is the interaction history and $a_t$ is an action selected by the model. One execution produces an execution trajectory $\tau_i=(o_{i,0},a_{i,0},\ldots,o_{i,T})$. The environment then exposes the benchmark-native performance signal directly as a scalar reward $r_i\in[0,1]$, instantiated as task success or test pass rate, together with verifier diagnostics $\mathbf{v}_i$, such as per-test outcomes and error messages. The complete environment transition is
\begin{equation}
    (\tau_i,r_i,\mathbf{v}_i)
    =\mathcal{E}(\mathbf{x},\mathcal{K}_i;\pi).
\end{equation}

During optimization, a reference trajectory $\tau^\star$ supplies privileged evidence about a successful solution process, but it is never provided as a runtime input to the compact model. Let $\mathbb{K}$ denote the space of natural-language skills, which includes optional task-local helper programs. For a fixed $\pi$, our objective is to find a skill that maximizes verifier performance,
\begin{equation}
    \begin{gathered}
        \mathcal{K}_{\mathbf{x}}^\star
        \in
        \arg\max_{\mathcal{K}\in\mathbb{K}}
        J_{\mathbf{x}}(\mathcal{K}), \\
        J_{\mathbf{x}}(\mathcal{K})
        =
        \mathbb{E}_{(\tau,r,\mathbf{v})\sim
        p_{\mathcal{E},\pi}(\cdot\mid\mathbf{x},\mathcal{K})}
        \bigl[r\bigr].
    \end{gathered}
\end{equation}
Here, $p_{\mathcal{E},\pi}(\cdot\mid\mathbf{x},\mathcal{K})$ denotes the joint distribution over trajectories, rewards, and verifier diagnostics induced by the environment and compact model when conditioned on $\mathbf{x}$ and $\mathcal{K}$; their realizations at optimization step $i$ are $(\tau_i,r_i,\mathbf{v}_i)$. Rather than differentiating through $\pi$ or $\mathcal{E}$, \method{} explores $\mathbb{K}$ through verifier-grounded, natural-language policy updates $\mathcal{K}_0,\mathcal{K}_1,\ldots,\mathcal{K}_I$.

\subsection{Natural-Language Skill Optimization}
\label{sec:natural-language-skill-optimization}

At each optimization step $i$, \method{} takes the fixed target instance $\mathbf{x}$ and the current task-specific skill $\mathcal{K}_i$ as input, as shown at the upper left of Figure~\ref{fig:framework}(a). At step $0$, it instantiates an initial skill $\mathcal{K}_0$ that captures a coarse execution workflow. Each skill serves as an executable language artifact that specifies a prescriptive procedure, concrete tool calls and paths, output contracts, and, when beneficial, task-local helper scripts. The system then traverses an iterative optimization loop connecting the environment, state, critic, replay memory, and actor to specialize this artifact for the small-scale LVLM.

\paragraph{Environment.}
The environment $\mathcal{E}$ wraps the benchmark-specific tool interface, workspace, and official verifier around the compact model $\pi$. Conditioned on $\mathcal{K}_i$, the model repeatedly observes the current workspace, selects a tool action, reads the resulting observation, and reflects before initiating its next action. The environment records this interaction as $\tau_i$ and invokes the official verifier to obtain the scalar reward $r_i$ and diagnostics $\mathbf{v}_i$. Thus, despite differences in tools and verifiers across benchmarks, each adapter implements the same transition
\begin{equation}
    \mathcal{E}:(\mathbf{x},\mathcal{K}_i;\pi)
    \longmapsto (\tau_i,r_i,\mathbf{v}_i).
\end{equation}

\paragraph{State \& Reward.}
The controller combines the rollout with the optimization-side reference to form the structured state quadruple
\begin{equation}
    \mathbf{s}_i
    =
    \bigl(\mathbf{x},\tau_i,\tau^\star,\mathbf{v}_i\bigr).
\end{equation}
Here, $\mathbf{x}$ has no step subscript because the target instance remains fixed, whereas $\tau_i$ and $\mathbf{v}_i$ describe the behavior and diagnostic outcome induced by the current skill. Pairing the current trajectory with $\tau^\star$ exposes not only whether execution failed, but also where its action sequence first departed from a successful strategy; the verifier diagnostics $\mathbf{v}_i$ ground that comparison in the task's actual acceptance criterion. Alongside $\mathbf{s}_i$, the environment returns a separate scalar reward $r_i\in[0,1]$, which represents binary task success for pass/fail benchmarks and the normalized test pass rate when partial credit is available. Thus, $r_i$ quantifies how well the current skill performs, while $\mathbf{v}_i$ provides the diagnostic evidence needed to explain that outcome. The analysis and ablation of the state are provided in Appendix~A.

\paragraph{Critic.}
A frontier-model critic $\mathcal{C}_{\phi}$ evaluates the current skill from the state, scalar reward, and prior optimization evidence:
\begin{equation}
    \mathbf{g}_i
    =
    \mathcal{C}_{\phi}
    \bigl(\mathbf{s}_i,r_i,\mathcal{K}_i,\mathcal{M}_i\bigr),
\end{equation}
where $\mathbf{g}_i$ denotes natural-language skill-modification suggestions. The scalar $r_i$ indicates how well the current skill performed, while $\mathbf{v}_i$ explains the verifier outcome. By anchoring its judgment in both signals, the critic contrasts $\tau_i$ with $\tau^\star$ to locate the earliest causal divergence. It systematically distinguishes missing procedural guidance from tool misuse, output-contract violations, and non-actionable infrastructure failures. Furthermore, it identifies content that already supports successful behavior and converts the diagnosis into concrete, localized editing instructions. Consequently, the actor receives an evidence-backed repair plan rather than an unconstrained request to rewrite the entire skill. The prompts and ablation of the critic are provided in Appendix~B.

\paragraph{Replay Memory.}
The replay memory $\mathcal{M}_i$ is a compact textual history of completed optimization steps, not a store of raw token-level transitions. As highlighted in Figure~\ref{fig:framework}(a), it retains three forms of reusable evidence, namely failure signatures with verifier diagnostics, critic-summary history, and accepted edits with their observed outcomes. The current critic diagnosis is recorded before actor execution, and once the resulting skill is evaluated, the corresponding edit and outcome are permanently archived in memory. Relevant records are supplied to both $\mathcal{C}_{\phi}$ and $\mathcal{A}_{\theta}$. This memory structure discourages repeated failures, protects previously effective behavior, and provides explicit evidence for retaining or rolling back a modification following a performance regression.

\paragraph{Actor.}
Given the critic's suggestions, a frontier-model actor $\mathcal{A}_{\theta}$ produces a bounded skill update
\begin{equation}
    \begin{gathered}
        \Delta_i
        =
        \mathcal{A}_{\theta}
        \bigl(\mathbf{x},\mathcal{K}_i,\tau^\star,\mathbf{g}_i,\mathcal{M}_i\bigr), \\
        \mathcal{K}_{i+1}
        =
        \operatorname{Apply}(\mathcal{K}_i,\Delta_i).
    \end{gathered}
\end{equation}
The update $\Delta_i$ is realized through four explicit editing operations, namely \textsc{Insert}, \textsc{Replace}, \textsc{Create}, and \textsc{Delete}, applied over the skill bundle. The actor preserves content identified as effective, adds precise behavioral constraints, and may synthesize deterministic task-local helpers when a small-scale LVLM would struggle with a long or error-prone procedure. Reference evidence is distilled strictly into runtime-input-dependent guidance and is never introduced as a direct runtime dependency or a precomputed answer. The prompts and ablation of the actor are provided in Appendix~C.

\begin{table*}[t]
\centering
{\small
\renewcommand{\arraystretch}{1.2}
\begin{tabular}{c|c|l|ccccc}
\specialrule{1.0pt}{0pt}{0pt}
Base model & \multicolumn{2}{c|}{Method} & SkillsBench & SkillLearnBench & SWE-Skills-Bench & GAIA & EarthBench \\
\hline
\multirow[c]{7}{*}{Qwen3.5-9B}
  & -- & No-skill                   & 1.45          & 23.83          & 26.20          & 44.58         & 59.68 \\
\cline{2-8}
  & \multirow[c]{2}{*}{Closed-source}
       & Human-authored            & 10.14         & 30.00          & 52.00          & \na          & \na \\
  &    & Manus                     & 57.97         & 27.56          & 62.40          & 46.18         & 71.24 \\
\cline{2-8}
  & \multirow[c]{4}{*}{Open-source}
       & AutoSkill                 & 53.62         & 25.61          & 44.10          & 44.98         & 71.77 \\
  &    & EvoSkill                  & 50.72         & 24.22          & 45.90          & 48.19         & 62.90 \\
  &    & SkillX                    & 60.87         & 27.78          & 58.80          & \best{49.40} & 66.13 \\
  &    & \textbf{\method{} (ours)} & \best{73.91}  & \best{32.11}  & \best{82.80}  & \best{49.40} & \best{76.08} \\
\specialrule{0.7pt}{0pt}{0pt}
\multirow[c]{7}{*}{Qwen3.5-4B}
  & -- & No-skill                   & 0.00          & 24.50          & 17.60          & 40.16         & 52.69 \\
\cline{2-8}
  & \multirow[c]{2}{*}{Closed-source}
       & Human-authored            & 5.80          & 28.94          & 42.70          & \na          & \na \\
  &    & Manus                     & 36.23         & 31.17          & 53.40          & 41.37         & \best{71.51} \\
\cline{2-8}
  & \multirow[c]{4}{*}{Open-source}
       & AutoSkill                 & 31.88         & 24.33          & 35.50          & 40.56         & 66.13 \\
  &    & EvoSkill                  & 40.58         & 22.89          & 37.30          & 42.17         & 67.47 \\
  &    & SkillX                    & \best{43.48}  & 30.22          & 48.40          & \best{44.18} & 65.86 \\
  &    & \textbf{\method{} (ours)} & 42.03         & \best{33.00}  & \best{66.70}  & 43.78         & \best{71.51} \\
\specialrule{1.0pt}{0pt}{0pt}
\end{tabular}
}
\caption{Main results on five benchmarks. All scores are three-run average score, accuracy or pass rate (\%). The best result in each model--benchmark group is shown in bold.}
\label{tab:main}
\end{table*}

\method{} repeats this process for $i=0,\ldots,I-1$, yielding the final composed policy $\mathcal{K}_I=\operatorname{Apply}(\cdots\operatorname{Apply}(\mathcal{K}_0,\Delta_0),\ldots,\Delta_{I-1})$. It also supports batch optimization. For $\mathcal{B}=\{\mathbf{x}^{(n)}\}_{n=1}^{N}$, parallel environment executions form $\mathbf{S}_i^{\mathcal{B}}=\{\mathbf{s}_i^{(n)}\}_{n=1}^{N}$ and the empirical verifier objective $\widehat{J}_i(\mathcal{B})=\frac{1}{N}\sum_{n=1}^{N}r_i^{(n)}$. Skills may be updated independently as $\{\mathcal{K}_i^{(n)}\}_{n=1}^{N}$ or tied across related instances by setting $\mathcal{K}_i^{(n)}=\mathcal{K}_i$ and aggregating their evidence before one update. As optimization advances, successive updates $\Delta_i$ incorporate progressively finer constraints, including input grounding, task-local computation, and self-validation, that effectively narrow the error-prone behaviors of the small-scale LVLM. Figure~\ref{fig:framework}(b) illustrates this evolution from a generic workflow to a grounded, executable, and self-correcting skill, with increasing verifier performance across steps.

\section{Experiments and Analysis}

In this section, we present comprehensive empirical comparisons against existing skill-generation methods across five benchmark families, analyze the dynamic evolution process of \method{}, examine the structural differences between automatically generated and human-authored skills, and evaluate the generation costs of various approaches. Detailed ablation studies of our framework are provided in Appendices~A-C.

\subsection{Experimental Setup}

\paragraph{Benchmarks.}
We evaluate our method across five diverse benchmark families. SkillsBench measures whether agents can use task-paired skills across diverse domains \citep{li2026skillsbench}, from which we select a subset of 26 tasks where each instance is resolved using a single skill. SWE-Skills-Bench evaluates skill utility on software-engineering tasks with execution-based tests \citep{han2026sweskills}, where we report performance on 117 instances covered by 10 high-difficulty skills. SkillLearnBench evaluates continual skill generation across 100 verified task instances \citep{zhong2026skilllearnbench}. GAIA covers multi-step information-seeking tasks \citep{mialon2023gaia}, where we evaluate on 165 tasks from the validation set. EarthBench covers Earth-science and data-processing workflows \citep{feng2025earthagent}, where we conduct testing across 248 evaluation samples. Specifically, GAIA and EarthBench are each treated as a single task where half of the instances are used for generating a skill, allowing us to report additional zero-shot test performance on the remaining samples, whereas for the remaining three benchmarks, a single instance from each task is used to generate the corresponding skill before evaluating across all instances of that task. Further dataset details and evaluation settings are provided in Appendix~D.

\paragraph{Models and baseline methods.}
The target small-scale LVLMs evaluated in our experiments are Qwen3.5-9B and Qwen3.5-4B \citep{qwen35blog}, which operate within the skill-executing agent loop of OpenCode \citep{opencode2026docs}. A strong frontier model, specifically GPT-5.4 \citep{singh2026openaigpt5card}, is utilized exclusively during the offline skill-generation phase to serve as the actor, and critic, and all downstream evaluation tokens consumed by the compact models are excluded from the reported generation costs. We compare our framework against standard no-skill execution, three open-source automated skill evolution methods consisting of AutoSkill \citep{yang2026autoskill}, EvoSkill \citep{alzubi2026evoskill}, and SkillX \citep{wang2026skillx}, as well as skills generated by the closed-source Manus system \citep{manus2026skills}. Human-authored skill baselines are reported only when officially provided by the benchmark \citep{li2026skillsbench,han2026sweskills,zhong2026skilllearnbench}. All reported results represent the average performance across three executions of the same skill, and \method{} is configured with a five-step reinforcement learning schedule across all tasks. Further baseline methods' details are provided in Appendix~E.

\subsection{Main Results and Analysis}
\label{sec:main-results-and-analysis}

\begin{table}[t]
\centering
{\small
\renewcommand{\arraystretch}{1.2}
\begin{tabular}{c|l|cc}
\specialrule{1.0pt}{0pt}{0pt}
Base model & Method & GAIA & EarthBench \\
\hline
\multirow[c]{6}{*}{Qwen3.5-9B}
  & No-skill                   & 37.00         & 58.87 \\
  & Manus                     & 45.53         & 70.43 \\
\cline{2-4}
  & AutoSkill                 & 47.97         & 70.16 \\
  & EvoSkill                  & 45.12         & 66.94 \\
  & SkillX                    & 42.28         & 67.20 \\
  & \textbf{\method{} (ours)} & \best{49.59} & \best{72.31} \\
\specialrule{0.7pt}{0pt}{0pt}
\multirow[c]{6}{*}{Qwen3.5-4B}
  & No-skill                   & 34.55         & 50.54 \\
  & Manus                     & 33.33         & \best{70.97} \\
\cline{2-4}
  & AutoSkill                 & 33.74         & 69.89 \\
  & EvoSkill                  & 34.15         & 66.67 \\
  & SkillX                    & 28.86         & 65.59 \\
  & \textbf{\method{} (ours)} & \best{40.65} & 69.62 \\
\specialrule{1.0pt}{0pt}{0pt}
\end{tabular}
}
\caption{Zero-shot results on the rest of GAIA and EarthBench. Scores are three-run average score and accuracy (\%).}
\label{tab:zero-shot}
\end{table}

As reported in Table~\ref{tab:main}, \method{} achieves superior overall performance across both model scales, demonstrating that natural-language reinforcement learning effectively adapts procedural knowledge to small-scale LVLMs. Notably, human-authored skills and generic skills generated by systems such as Manus frequently yield suboptimal gains or even degrade performance compared to standard automated evolution baselines. This phenomenon directly confirms our core motivation regarding the model-mismatch bottleneck. Skills composed by human experts or strong frontier models implicitly assume expansive reasoning capacities, broad context tolerance, and robust error recovery. When these high-level instructions are injected into a compact model, they frequently induce cognitive overload and trigger severe argument hallucinations. In contrast, by treating the target compact model as the interactive environment, \method{} diagnoses specific execution failures and injects localized, prescriptive boundaries that small-scale LVLMs can reliably execute.

The advantages of executor-specific policy optimization are most pronounced on benchmarks characterized by complex procedural workflows and strict tool-use conventions, such as SWE-Skills-Bench and SkillsBench. On SWE-Skills-Bench, \method{} outperforms all open-source and closed-source baselines by substantial margins on Qwen3.5-9B and achieves a dominating pass rate on Qwen3.5-4B. Software engineering tasks strictly penalize unverified file edits, out-of-order tool invocations, and ungrounded argument fabrication. Generic skill generators frequently produce high-level advice that fails to prevent these mechanical errors. Our critic-actor loop directly repairs these failure modes by inserting mandatory verification steps and input-grounding constraints, enabling compact models to execute multi-step engineering pipelines with unprecedented precision.

On benchmarks requiring open-ended information seeking, continual task adaptation, and complex domain workflows, including SkillLearnBench, GAIA, and EarthBench, \method{} consistently matches or surpasses the strongest baseline methods. While procedural constraints cannot substitute for missing factual knowledge or complex mathematical reasoning during multi-hop retrieval, our learned skills effectively regulate the search process, enforce systematic evidence verification, and prevent premature task finalization. Consequently, \method{} secures leading accuracy on SkillLearnBench and EarthBench while matching the top-performing baseline on GAIA. These results confirm that natural-language policy optimization provides reliable behavioral control even when task success heavily depends on external data interpretation.

Perhaps most remarkably, the empirical analysis reveals that executor-specific skill optimization can effectively bridge substantial model capacity gaps. As shown in Table~\ref{tab:main}, Qwen3.5-4B equipped with \method{} achieves a pass rate on SWE-Skills-Bench that surpasses the performance of the larger Qwen3.5-9B model when deployed with human-authored skills, AutoSkill, EvoSkill, SkillX, or even Manus-generated skills. This finding demonstrates that for structured real-world tasks, an optimized natural-language policy that constrains error-prone tendencies is more valuable than raw parameter scaling. By tailoring procedural control to the exact action distribution of the underlying small-scale LVLM, our framework enables lightweight compact models to outperform unoptimized models of more than twice their parameter scale, establishing a compelling paradigm for cost-efficient agent deployment.

\subsection{Zero-Shot Results and Analysis}

\begin{figure}[t]
  \centering
  \begin{subfigure}[t]{0.49\linewidth}
    \centering
    \includegraphics[width=\linewidth]{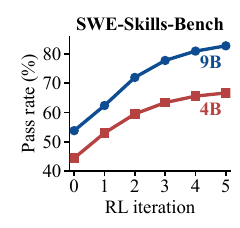}
    \caption{}
  \end{subfigure}\hfill
  \begin{subfigure}[t]{0.49\linewidth}
    \centering
    \includegraphics[width=\linewidth]{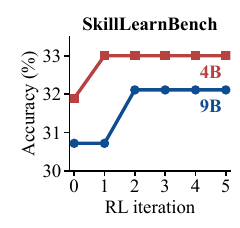}
    \caption{}
  \end{subfigure}
  \caption{Learning dynamics through five SKILLER iterations. The vertical axis is the three-repeat average pass rate and accuracy.}
  \label{fig:rl_trends}
\end{figure}

To evaluate the out-of-distribution generalization of our learned policies, Table~\ref{tab:zero-shot} reports the zero-shot transfer performance on the held-out half of GAIA and EarthBench. On Qwen3.5-9B, \method{} outperforms all baselines by substantial margins, indicating that our optimization loop extracts truly reusable procedural rules rather than overfitting to the surface features of the skill-generation instances. By enforcing systematic evidence gathering and strict output normalization, the generated skills provide robust behavioral scaffolding for small-scale LVLMs across unseen problem distributions. Under tighter capacity constraints on Qwen3.5-4B, \method{} maintains remarkable resilience by securing the highest accuracy on GAIA and remaining highly competitive on EarthBench. Notably, while closed-source Manus skills slightly lead on the EarthBench evaluation, they severely degrade performance on GAIA, where both Manus and SkillX fall below the standard no-skill baseline. This divergence highlights a fundamental trade-off where verbose domain context frequently triggers cognitive overload and error propagation during complex multi-hop reasoning. In contrast, \method{} constructs concise and executor-specific procedural boundaries that consistently mitigate hallucinations, confirming that lightweight compact models benefit most from disciplined behavioral control rather than exhaustive context injection.

\subsection{Learning Dynamics}
\label{sec:learning-dynamics}

To understand how natural-language reinforcement learning iteratively refines policy artifacts over time, Figure~\ref{fig:rl_trends} illustrates the optimization trajectories across five steps on SWE-Skills-Bench and SkillLearnBench. The contrasting curves reveal that the pace of policy convergence is fundamentally governed by the procedural complexity of the target domain. As shown in Figure~\ref{fig:rl_trends}(a), both small-scale LVLMs exhibit substantial and continuous performance gains across all five iterations on SWE-Skills-Bench, demonstrating that complex software engineering workflows benefit from cumulative policy specialization where early steps resolve coarse execution failures and later iterations inject finer constraints like input grounding and self-validation. Conversely, on SkillLearnBench in Figure~\ref{fig:rl_trends}(b), both compact models achieve rapid convergence within the first two optimization steps, indicating that when task bottlenecks stem primarily from output contract formatting or basic tool routing, our critic module quickly isolates the causal errors to establish an effective behavioral boundary.

\begin{table}[t]
  \centering
  {\small
  \renewcommand{\arraystretch}{1.2}
  \begin{tabular}{l|ccccc}
    \specialrule{1.0pt}{0pt}{0pt}
    Method & Words & TF-IDF & Scripts & LOC & Refs. \\
    \hline
    Human & 1162.52 & 0.07 & 1.48 & 8,819 & 0.43 \\
    Manus & 346.57 & 0.06 & 2.42 & 11,168 & 0.34 \\
    AutoSkill & 1887.41 & 0.93 & 0.54 & 10,527 & 0.22 \\
    EvoSkill & 1256.98 & 0.89 & 1.69 & 7,236 & 0.45 \\
    SkillX & 427.48 & 0.15 & 2.06 & 14,701 & 0.43 \\
    \method{} & 534.33 & 0.07 & 2.96 & 15,747 & 0.47 \\
    \specialrule{1.0pt}{0pt}{0pt}
  \end{tabular}
  }
  \caption{Structural statistics of generated skills on SkillsBench. Evaluated metrics include natural-language verbosity denoted by Words, inter-task instruction diversity measured via TF-IDF cosine similarity, mean script and references count per task, total physical lines of code denoted by LOC.}
  \label{tab:human-skiller-skill-stats}
\end{table}

\subsection{Structural Analysis of Generated Skills}
\label{sec:structural-analysis}

Table~\ref{tab:human-skiller-skill-stats} details structural statistics on SkillsBench to uncover fundamental differences between human-authored and automated policies. Open-source baselines consistently produce verbose prompts with elevated TF-IDF similarity, indicating a reliance on repetitive boilerplate templates that induce cognitive overload for compact models. In stark contrast, \method{} yields concise instructions with remarkably low TF-IDF scores matching human developers, proving our optimization loop synthesizes highly specialized behavioral constraints rather than generic patterns. Furthermore, \method{} registers the highest number of discrete scripts and total lines of code. This distribution highlights a deliberate paradigm shift that offloads complex procedural reasoning from natural language into deterministic external tools. By minimizing prompt verbosity while maximizing code-level execution capabilities, the generated skills perfectly accommodate the limited context windows of small-scale LVLMs. Generation examples are provided in Appendix~F.

\begin{table}[t]
\centering
{\small
\renewcommand{\arraystretch}{1.2}
\begin{tabular}{ccccc}
\toprule
Method & Input & Output & Cost (\$) & Avg. Score \\
\midrule
AutoSkill & 0.86M & 0.02M & 2.53 & 48.02 \\
EvoSkill & 0.34M & 0.07M & 1.95 & 46.39 \\
SkillX & 3.61M & 0.37M & 14.55 & 52.60 \\
\method{} & 2.68M & 0.15M & 8.95 & \best{62.86} \\
\bottomrule
\end{tabular}
}
\caption{Cost analysis and average performance of skill generation on Qwen3.5-9B across five benchmarks. Reported metrics include cumulative strong-model input and output token consumption in millions, monetary cost by GPT-5.4, and the average performance.}
\label{tab:cost}
\end{table}

\subsection{Cost-Effectiveness Analysis}
\label{sec:cost-effectiveness}

Table~\ref{tab:cost} presents a cost-effectiveness analysis alongside average performance on Qwen3.5-9B, revealing a fundamental trade-off between optimization depth and computational expenditure. While baselines like AutoSkill incur minimal financial costs, their shallow prompt rewriting fails to resolve the model-mismatch bottleneck and yields constrained accuracy. Conversely, SkillX consumes an excessive volume of output tokens and incurs the highest monetary penalty, indicating that open-ended generation without precise environment grounding causes inefficient textual bloat. \method{} optimally resolves this trade-off by utilizing precise execution feedback to drive highly targeted policy updates. By restricting unnecessary text generation steps, our framework achieves a commanding performance advantage while remaining substantially more cost-efficient than exhaustive generation methods, proving that targeted behavioral alignment offers the highest return on investment for empowering small-scale LVLMs. The cost details of each benchmark and steps are provided in Appendix~G.

\section{Conclusion}

In this work, we introduced \method{}, a natural-language-driven reinforcement learning framework designed to resolve the model-mismatch bottleneck and automatically generate executor-specific skills for small-scale LVLMs. By treating the target compact model's agent loop as an interactive environment and propagating all diagnostic signals entirely through structured text, our framework iteratively refines textual policies based on precise execution feedback without requiring neural weight updates. This language-level policy iteration successfully offloads complex procedural reasoning into deterministic external tools, thereby establishing robust behavioral boundaries that prevent cognitive overload and mechanical errors. Extensive empirical evaluations across five diverse benchmarks demonstrate that \method{} consistently outperforms both open-source skill evolution methods and closed-source generation systems. Most remarkably, these tailored behavioral constraints enable lightweight compact models to surpass the task-specific performance of unoptimized models at more than twice their parameter scale, establishing a highly cost-effective and scalable paradigm for real-world agent deployment.

\clearpage
\bibliography{references}

\clearpage
\suppressfloats[t]
\section*{Appendix A: State Analysis and Ablation}
\label{app:state-ablation}

We examine the contribution of each component in the structured state
$\mathbf{s}_i=(\mathbf{x},\tau_i,\tau^\star,\mathbf{v}_i)$ using
Qwen3.5-9B. Each ablation removes one component while retaining the remaining
optimization and evaluation settings. Table~\ref{tab:state-ablation} reports
average performance over three runs under the official metric of each
benchmark.

The ablation reveals a clear asymmetry between components that ground a policy
update and those that refine it. Removing either the task instance
$\mathbf{x}$ or the current trajectory $\tau_i$ causes substantially greater
damage than removing $\tau^\star$ or $\mathbf{v}_i$, particularly on
SkillsBench and SWE-Skills-Bench. A successful reference or verifier outcome
cannot by itself identify a useful edit: $\mathbf{x}$ specifies what the
executor must satisfy, whereas $\tau_i$ exposes how the current skill actually
shapes its behavior. Together, they turn feedback into an error signal that is
conditioned on the task rather than generic advice.

The dependence on this grounding pair is strongest for benchmarks with strict
tool sequences, executable artifacts, and output contracts. GAIA degrades less
sharply under every ablation. This pattern suggests that information seeking
tasks can retain partial utility from broadly applicable search procedures even
when one state signal is absent. In contrast, structured skill and software
tasks require the editor to connect a precise failure to a precise action or
artifact, which makes omissions from the state much harder to compensate for.

The remaining two components play complementary refinement roles. Across all
three benchmarks, removing the reference trajectory is more harmful than
removing verifier diagnostics. This result indicates that evidence about the
execution process offers a richer target for localized edits than outcome
feedback alone. Nevertheless, the consistent loss without $\mathbf{v}_i$ shows
that a plausible execution path is insufficient unless it is anchored to the
benchmark's acceptance criterion. The full state is effective because it
combines task intent, observed behavior, a positive execution reference, and
correctness grounded in verifier feedback rather than relying on any single
feedback source.

\begin{table}[t]
  \centering
  {\small
  \renewcommand{\arraystretch}{1.2}
  \begin{tabular}{@{}lccc@{}}
    \toprule
    Variant & SkillsBench & \shortstack{SWE-Skills-\\Bench} & GAIA \\
    \midrule
    \textbf{\method{}} & \textbf{73.91} & \textbf{82.80} & \textbf{49.40} \\
    w/o $\mathbf{x}$ & 1.45 & 20.51 & 39.76 \\
    w/o $\tau_i$ & 2.44 & 24.87 & 44.58 \\
    w/o $\tau^\star$ & 58.71 & 62.16 & 44.98 \\
    w/o $\mathbf{v}_i$ & 60.87 & 64.50 & 46.20 \\
    \bottomrule
  \end{tabular}
  }
  \caption{State component ablation using Qwen3.5-9B as the executor. Scores are means over three runs in percentage points under each benchmark's official metric, with higher values indicating better performance.}
  \label{tab:state-ablation}
\end{table}

\section*{Appendix B: Critic Prompt Analysis and Ablation}
\label{app:critic-ablation}

The critic prompt implements four operations that mirror the critic module in
Figure~1 of the main paper. It evaluates the current execution, compares the
observed and reference trajectories, locates the earliest causal error, and
generates a bounded skill modification. We remove each operation separately
while retaining the remaining prompt instructions and experimental settings.
Table~\ref{tab:critic-ablation} reports performance with Qwen3.5-9B.

\begin{table}[ht]
  \centering
  {\small
  \renewcommand{\arraystretch}{1.2}
  \begin{tabular}{@{}lccc@{}}
    \toprule
    Variant & SkillsBench & \shortstack{SWE-Skills-\\Bench} & GAIA \\
    \midrule
    \textbf{\method{}} & \textbf{73.91} & \textbf{82.80} & \textbf{49.40} \\
    w/o Evaluate & 60.87 & 68.20 & 46.40 \\
    w/o Compare & 60.87 & 64.50 & 46.40 \\
    w/o Locate Error & 58.71 & 64.46 & 45.00 \\
    w/o Generation & 36.44 & 24.90 & 42.15 \\
    \bottomrule
  \end{tabular}
  }
  \caption{Critic prompt ablation using Qwen3.5-9B as the executor. Each variant removes one critic operation, and scores follow the official metric of each benchmark, with higher values indicating better performance.}
  \label{tab:critic-ablation}
\end{table}

The dominant failure after removing Generation suggests that translation is the
critic's decisive function. Evaluation, comparison, and localization can expose
a defect, but the learning loop improves only when the diagnosis becomes a
bounded edit to the textual policy. Without this operation, rollout evidence
remains descriptive and cannot alter the behavior that produced the failure.

Comparison and error localization provide a second layer of control. The
official verifier already supplies a coarse outcome signal, which may partially
compensate for the absence of an explicit evaluation instruction. It cannot,
however, explain which decision caused the outcome. Comparing the observed path
with a successful reference narrows the search to meaningful divergences, and
localizing the earliest error discourages the actor from patching downstream
symptoms. This distinction is especially important for software tasks, where a
late test failure can originate from an earlier choice of file, tool, or
verification procedure.

The smaller sensitivity on GAIA suggests a boundary of prompt based skill
repair. One plausible explanation is that information seeking tasks may fail
because relevant external evidence was not found, rather than because a visible
procedure was applied incorrectly. A critic can reshape the search policy, but
it cannot supply missing facts. Structured skill and software tasks expose more
repeatable causal traces, which give comparison and localization greater
leverage. The complete prompt is effective because it composes a verdict, a
contrastive reference, a causal diagnosis, and an executable update into one
feedback path.

\begin{figure*}[p]
  \centering
  \includegraphics[width=\textwidth,height=0.88\textheight,keepaspectratio]{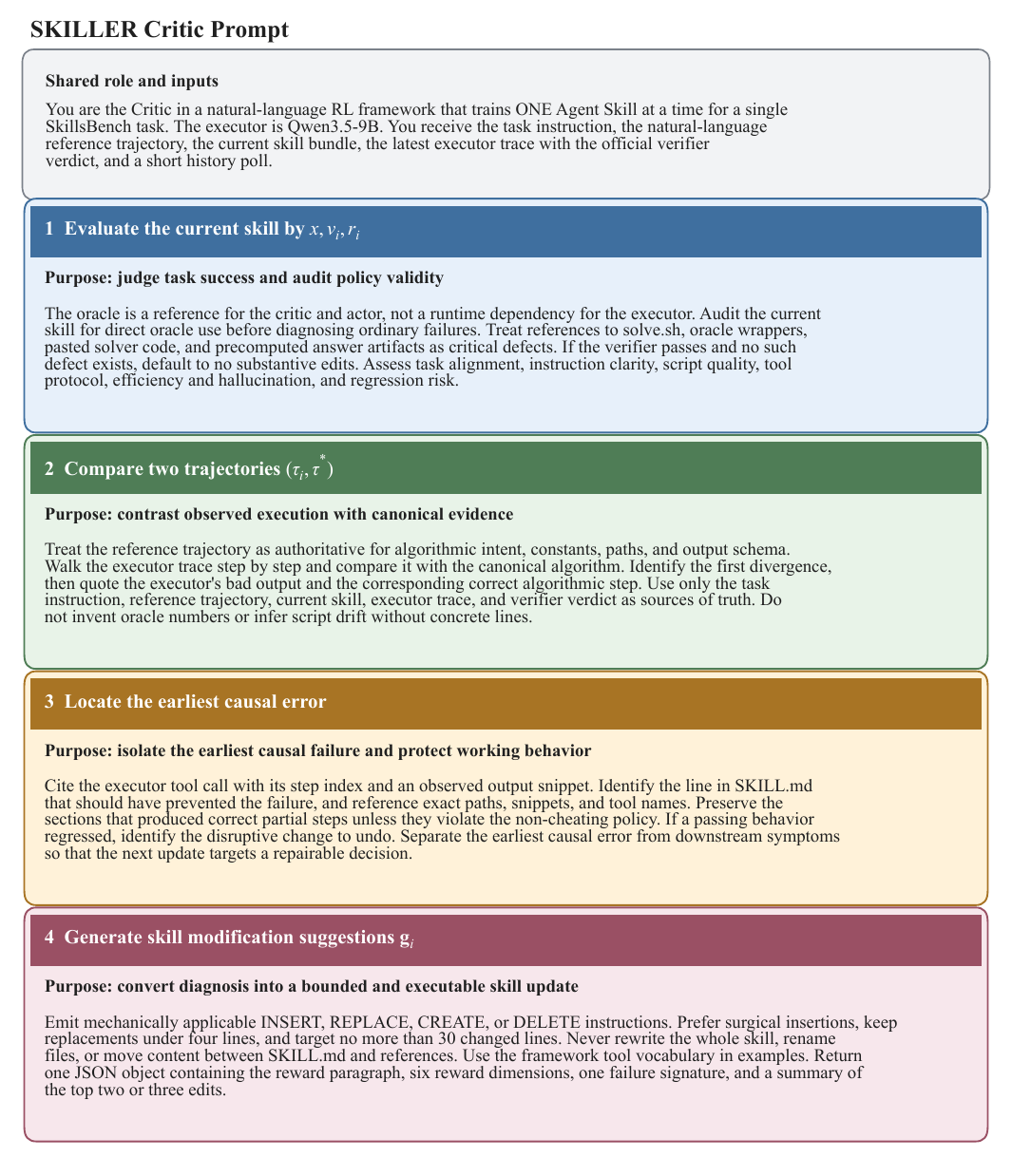}
  \caption{SKILLER critic prompt used for SkillsBench and organized by its four operations. Each colored panel uses a distinct visual treatment and begins with the purpose of the corresponding operation. The prompt content is drawn from the released implementation and grouped to remove repeated instructions while preserving the operational constraints.}
  \label{fig:skillsbench-critic-prompt}
\end{figure*}

Figure~\ref{fig:skillsbench-critic-prompt} presents the SKILLER critic prompt
used for SkillsBench using the four operations examined by the method and
ablation.

\section*{Appendix C: Actor Prompt Analysis and Ablation}
\label{app:actor-ablation}

The actor prompt implements four operations that mirror the actor module in
Figure~1 of the main paper. It applies bounded edit operations, synthesizes task
local helper scripts, preserves effective skill content, and emits a complete
skill update. We ablate the first three operations separately while retaining
the remaining prompt instructions and experimental settings. The final
operation defines the update interface itself, so removing it would disable the
learning step rather than isolate a comparable prompt component.
Table~\ref{tab:actor-ablation} reports performance with Qwen3.5-9B.

\begin{table}[ht]
  \centering
  {\small
  \renewcommand{\arraystretch}{1.2}
  \begin{tabular}{@{}lccc@{}}
    \toprule
    Variant & SkillsBench & \shortstack{SWE-Skills-\\Bench} & GAIA \\
    \midrule
    \textbf{\method{}} & \textbf{73.91} & \textbf{82.80} & \textbf{49.40} \\
    w/o Operations & 68.45 & 77.90 & 44.98 \\
    w/o Scripts & 55.34 & 52.30 & 43.68 \\
    w/o Preserve & 59.18 & 68.20 & 42.59 \\
    \bottomrule
  \end{tabular}
  }
  \caption{Actor prompt ablation using Qwen3.5-9B as the executor. Each variant removes one actor operation, and scores follow the official metric of each benchmark, with higher values indicating better performance.}
  \label{tab:actor-ablation}
\end{table}

The strongest degradation on SkillsBench and SWE-Skills-Bench occurs when the
actor cannot synthesize helper scripts. This pattern identifies executable
abstraction as a central bridge between language feedback and reliable action.
Long procedures embedded directly in a skill remain vulnerable to truncation,
format drift, and inconsistent tool use by a small executor. A task local
helper moves deterministic computation into a reusable artifact, leaving the
skill to specify when and how that artifact should be invoked. The larger effect
on structured skill and software tasks is consistent with their strict output
contracts and repeatable computation paths.

Preservation governs a different failure mode. A critic observes one rollout,
so its repair signal is necessarily local to the latest error. Without an
explicit instruction to retain effective content, the actor can overfit that
signal by replacing instructions that supported previously correct behavior.
This stability constraint is particularly important on GAIA, where broadly
useful search and verification routines must survive updates driven by
heterogeneous tasks. Preservation therefore protects accumulated competence
that is not visible in the current trajectory.

Bounded operations yield a smaller but consistent contribution. Their role is
to control the scope of plasticity rather than introduce new procedural
knowledge. Exact insertion, replacement, creation, and deletion primitives
make critic feedback mechanically actionable and reduce unintended edits to
unrelated files. Together, the three operations form a coherent update policy.
Helper synthesis adds executable competence, preservation retains established
competence, and bounded editing mediates the tradeoff between the two.

\begin{figure*}[p]
  \centering
  \includegraphics[width=\textwidth,height=0.88\textheight,keepaspectratio]{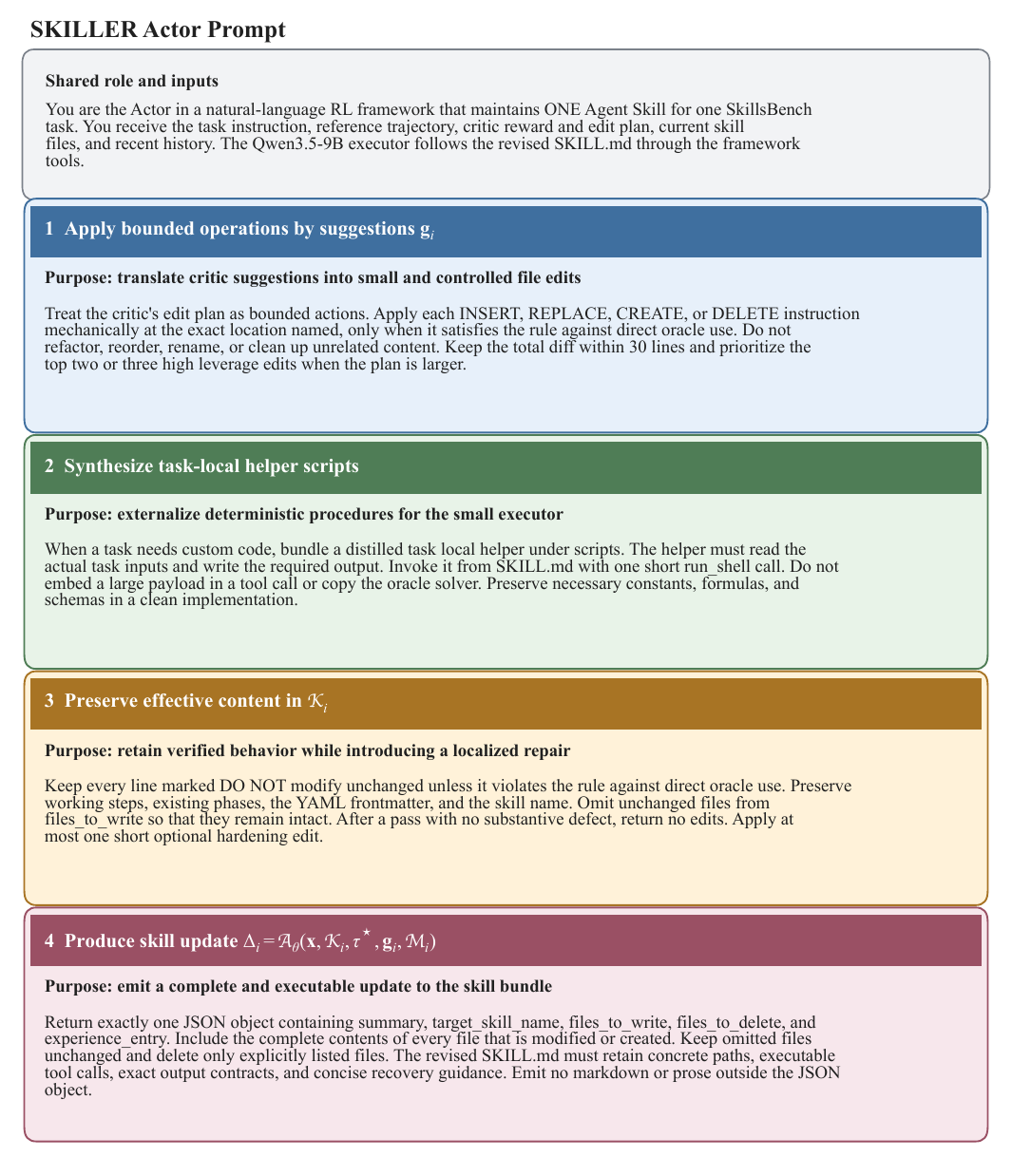}
  \caption{SKILLER actor prompt used for SkillsBench and organized by its four operations. Each colored panel uses a distinct visual treatment and begins with the purpose of the corresponding operation. The prompt content is drawn from the released implementation and grouped to remove repeated instructions while preserving the operational constraints.}
  \label{fig:skillsbench-actor-prompt}
\end{figure*}

Figure~\ref{fig:skillsbench-actor-prompt} presents the SKILLER actor prompt used
for SkillsBench using the four operations represented by the method.

\section*{Appendix D: Benchmark Task Selection and Data Splits}
\label{app:benchmark-splits}

We use \emph{task} to denote a benchmark unit associated with a skill and
\emph{instance} to denote an individual sample evaluated under that task. The
SkillsBench subset contains 26 tasks, each selected because it can be solved
with a single skill. For SWE-Skills-Bench, we retain the tasks for which the
original benchmark reports that adding a skill changes performance, including
both gains and declines. This criterion yields 10 tasks and 117 instances. The
subset is designed to study settings in which skill use has a measurable
behavioral effect, and its results should not be interpreted as estimates over
every task in the original benchmark. SkillLearnBench uses the official
configuration and verifier.

GAIA and EarthBench are each treated as one task, and each dataset sample is
treated as an instance. For GAIA, we stratify the samples by the three
difficulty levels and randomly assign half of each stratum to skill generation.
The resulting generation split contains 83 instances. The remaining instances
form the zero-shot test split. For EarthBench, we stratify the samples by the
three task categories and apply the same procedure. The generation split
contains 124 instances, and the remaining half forms the zero-shot test split.

\begin{table*}[t]
  \centering
  {\small
  \renewcommand{\arraystretch}{1.2}
  \begin{tabular}{lrrrrrrrr}
    \toprule
    Method or stage & R1 & R2 & R3 & Avg. & Acc. (\%) & Input (K) & Output (K) & Total (K) \\
    \midrule
    AutoSkill       & 28 & 27 & 23 & 26.00 & 25.61 & 1352.98 & 30.31 & 1383.29 \\
    EvoSkill        & 23 & 25 & 24 & 24.00 & 24.22 & 366.65  & 21.62 & 388.27 \\
    SkillX          & 26 & 27 & 28 & 27.00 & 27.78 & 754.82  & 104.16 & 858.98 \\
    \midrule
    \method{} Step 1 & 30 & 27 & 31 & 29.33 & 30.33 & 162.22 & 73.66 & 235.88 \\
    \method{} Step 2 & 31 & 30 & 31 & 30.67 & \textbf{32.11} & 185.34 & 73.93 & 259.27 \\
    \method{} Step 3 & 28 & 25 & 33 & 28.67 & 30.17 & 184.98 & 73.92 & 258.90 \\
    \method{} Step 4 & 31 & 30 & 27 & 29.33 & 30.06 & 205.06 & 83.82 & 288.88 \\
    \method{} Step 5 & 29 & 29 & 31 & 29.67 & 30.17 & 180.16 & 84.67 & 264.83 \\
    \bottomrule
  \end{tabular}
  }
  \caption{Detailed SkillLearnBench results and skill generation tokens using Qwen3.5-9B as the executor. R1, R2, and R3 are the numbers of passed instances out of 100 in three repeated runs, and Avg. is their mean. Acc. is the official accuracy averaged over the same runs and expressed as a percentage. Input, Output, and Total report strong model token consumption in thousands for each baseline or SKILLER stage.}
  \label{tab:skilllearnbench-cost}
\end{table*}

Stratification preserves the benchmark composition across the generation and
test partitions. No held-out instance is used to update the skill, so the
zero-shot results measure whether a skill induced from one subset transfers to
unseen instances from the same benchmark.

\section*{Appendix E: Method Implementation Details}
\label{app:implementation-details}

The implementation assigns each task an independent skill bundle and
experience buffer, then initializes the bundle by refining the baseline skill
with the task instruction and reference trajectory while retaining unchanged
local files. The released SkillsBench configuration runs three optimization
rounds, limits each executor rollout to 30 tool steps, and uses sampling
temperatures of 0.2 for the executor and 0.3 for the actor and critic. During
each round, the executor runs the current skill in the task environment and the
benchmark verifier supplies the success signal, after which the critic
conditions on the task instruction, reference trajectory, current skill file
tree, rollout state, verifier result, and three most recent history records to
produce structured natural language feedback. The actor uses this feedback to
apply bounded edits and return the complete contents of modified files, and the
merged update is rejected if it directly uses oracle solver content. The system
stores skill snapshots before and after every round and restores the most
recent passing snapshot when a later update causes failure, which prevents
subsequent optimization from discarding an already successful skill.

\section*{Appendix F: Qualitative Skill Evolution}
\label{app:skill-evolution}

\begin{figure*}[p]
  \centering
  \includegraphics[width=\textwidth,height=0.88\textheight,keepaspectratio]{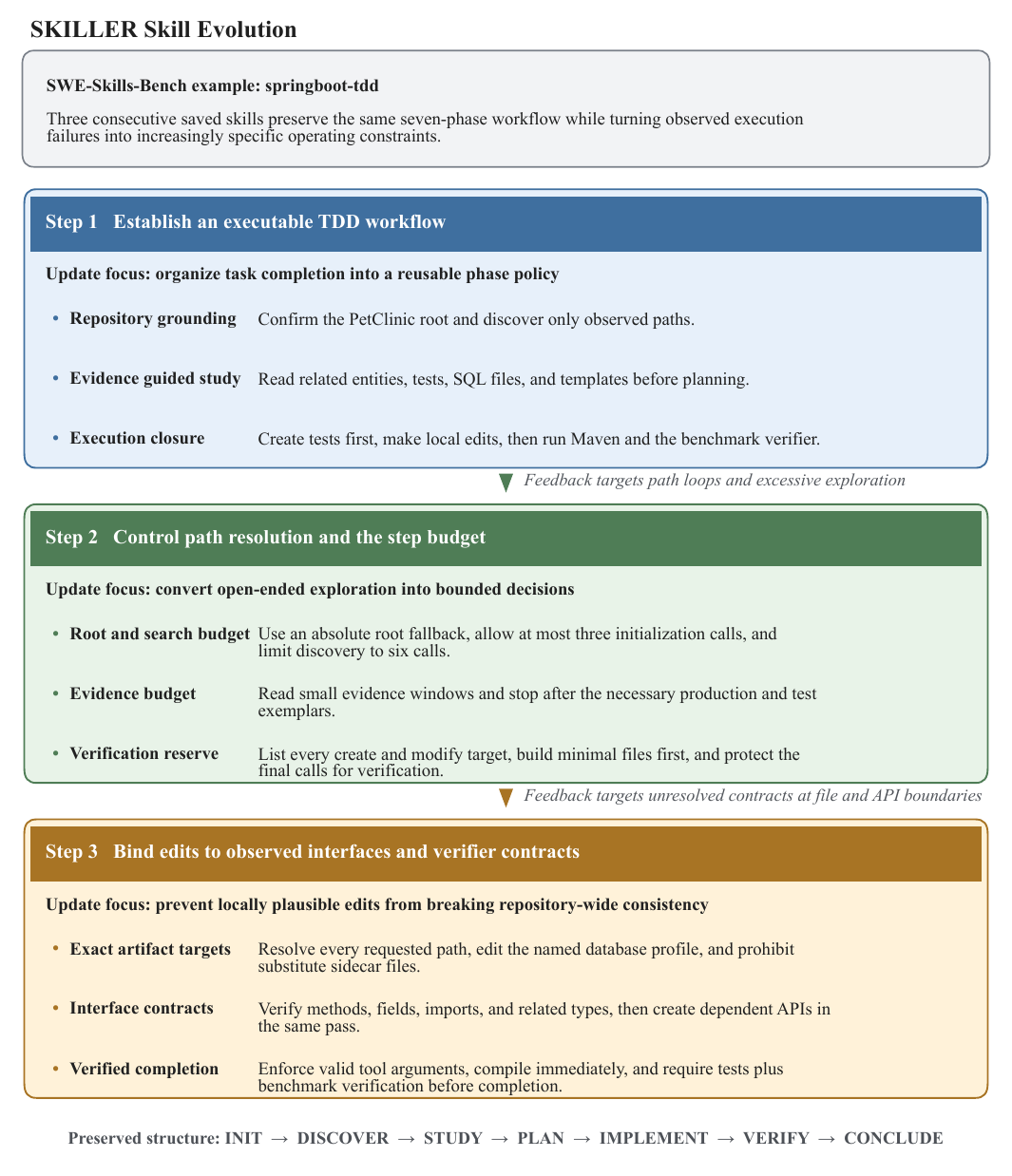}
  \caption{Evolution of the \texttt{springboot-tdd} skill on SWE-Skills-Bench across three SKILLER steps. Each panel summarizes the semantic changes introduced by one saved skill version rather than presenting a literal text comparison. The seven-phase workflow is retained while feedback progressively constrains path discovery, edit planning, artifact consistency, and verification.}
  \label{fig:swe-skill-evolution}
\end{figure*}

We examine three consecutive saved versions of the \texttt{springboot-tdd}
skill from SWE-Skills-Bench. This example was selected after comparing the
recorded skills from all five benchmarks because it contains substantive and
interpretable updates at both transitions while preserving the same overall
workflow. Figure~\ref{fig:swe-skill-evolution} summarizes the semantic changes
in the saved rounds, which we refer to as steps in this appendix.

The first transition changes the skill from a broad procedural guide into a
budget-aware control policy. Absolute root anchoring, finite exploration
limits, bounded evidence collection, and an explicit file checklist define
stopping conditions for actions that were previously open ended. These
changes reveal that an important source of failure is not missing domain
knowledge, but the allocation of a limited interaction budget. Encoding the
observed failure as a reusable constraint reduces the chance that a later
executor repeats the same navigation loop while leaving the task workflow
available for new instances.

The second transition moves the repair boundary from action selection to
artifact consistency. Exact database targets, observed method signatures,
cross-file dependencies, and valid tool arguments become preconditions for an
edit. This distinction matters because a locally plausible Java class can
still fail when its imports, repository methods, SQL profile, or tool call
does not match the surrounding project. The skill therefore converts
trajectory feedback into interface contracts that can prevent an entire
family of related failures rather than patching only the latest output.

Across all three steps, the seven execution phases remain unchanged while
their transition and completion criteria become more precise. This pattern
shows how bounded language-level updates can balance plasticity with
retention. New failure evidence is accumulated as local policy constraints,
whereas previously useful structure remains available to the executor. The
example is qualitative rather than a standalone measure of average behavior,
but it exposes the mechanism through which repeated interaction can produce a
more executable and regression-resistant skill.

\section*{Appendix G: SkillLearnBench Cost and Learning Dynamics}
\label{app:skilllearnbench-cost}

Table~\ref{tab:skilllearnbench-cost} reports the detailed SkillLearnBench
results for Qwen3.5-9B together with the strong model tokens used to generate
each skill or update. We omit No-skill and Human-authored because they do not
invoke a strong model for skill generation, and we omit Manus because its
token usage is unavailable.

The comparison separates generation volume from skill effectiveness.
AutoSkill consumes the most tokens among the reported baselines but does not
obtain the strongest baseline accuracy, while SkillX performs better with a
smaller token budget. Every reported SKILLER step also exceeds the strongest
baseline accuracy. This pattern indicates that generation volume alone does
not determine whether the resulting instructions provide useful behavioral
control for the target executor.

The SKILLER trajectory is nonmonotonic. Step 2 establishes the best checkpoint,
while later updates retain much of the gain but do not improve on that
checkpoint. This behavior is consistent with localized edits that repair a
sampled failure while occasionally narrowing instructions that remain useful
for other instances. It also motivates the snapshot and rollback mechanism
described in Appendix~E because the final update need not be the most
transferable policy.

The five update stages consume similar token budgets despite producing
different evaluation outcomes. In particular, the stage with the largest
token usage does not yield the highest accuracy. The marginal utility of an
update therefore depends more on the information contained in verifier
grounded feedback than on raw generation volume. Because the SKILLER rows
report stage level rather than cumulative token usage, this table characterizes
the efficiency of individual updates rather than the total expense of the full
optimization trajectory. The optimal stopping step can also vary with the
benchmark and executor.

\end{document}